\documentclass[final]{cvpr}

\usepackage{times}
\usepackage{epsfig}
\usepackage{subfiles}
\usepackage{graphicx}
\usepackage{amsmath}
\usepackage{amssymb}
\usepackage{comment}
\usepackage{enumitem}
\usepackage{dsfont}
\usepackage{booktabs}
\usepackage{graphics}
\usepackage{multirow}

\usepackage[pagebackref=true,breaklinks=true,colorlinks,bookmarks=false]{hyperref}

\def\cvprPaperID{5028} 
\def\confYear{CVPR 2022}

\begin{document}


\title{HAA4D: Few-Shot Human Atomic Action Recognition via \\ 3D Spatio-Temporal Skeletal Alignment}

\author{Mu-Ruei Tseng$^1$, Abhishek Gupta$^1$, Chi-Keung Tang$^1$, Yu-Wing Tai$^2$  \\
$^1$The Hong Kong University of Science and Technology, $^2$Kuaishou Technology \\
}

\maketitle

\begin{abstract}
Human actions involve complex pose variations and their 2D projections can be highly ambiguous. Thus 3D spatio-temporal or 4D (i.e., 3D+T) human skeletons, which are photometric and viewpoint invariant, are an excellent alternative to 2D+T skeletons/pixels to improve action recognition accuracy.  This paper proposes a new 4D dataset {\bf HAA4D} which consists of more than 3,300 RGB videos in 300 human atomic action classes. HAA4D is clean, diverse, class-balanced where each class is viewpoint-balanced with the use of 4D skeletons, in which as few as one 4D skeleton per class is sufficient for training a deep recognition model. Further, the choice of atomic actions makes annotation even easier, because each video clip lasts for only a few seconds. All training and testing 3D skeletons in HAA4D are globally aligned, using a deep alignment model to the same global space, making each skeleton face the negative $z$-direction. Such alignment makes matching skeletons more stable by reducing intraclass variations and thus with fewer training samples per class needed for action recognition.  Given the high diversity and skeletal alignment in HAA4D, we construct the first baseline few-shot 4D human atomic action recognition network without bells and whistles, which produces comparable or higher performance than relevant state-of-the-art techniques relying on embedded space encoding without explicit skeletal alignment, using the same small number of training samples of unseen classes.
\end{abstract}

\begin{figure}[t]
  \centering
  \includegraphics[scale=0.16]{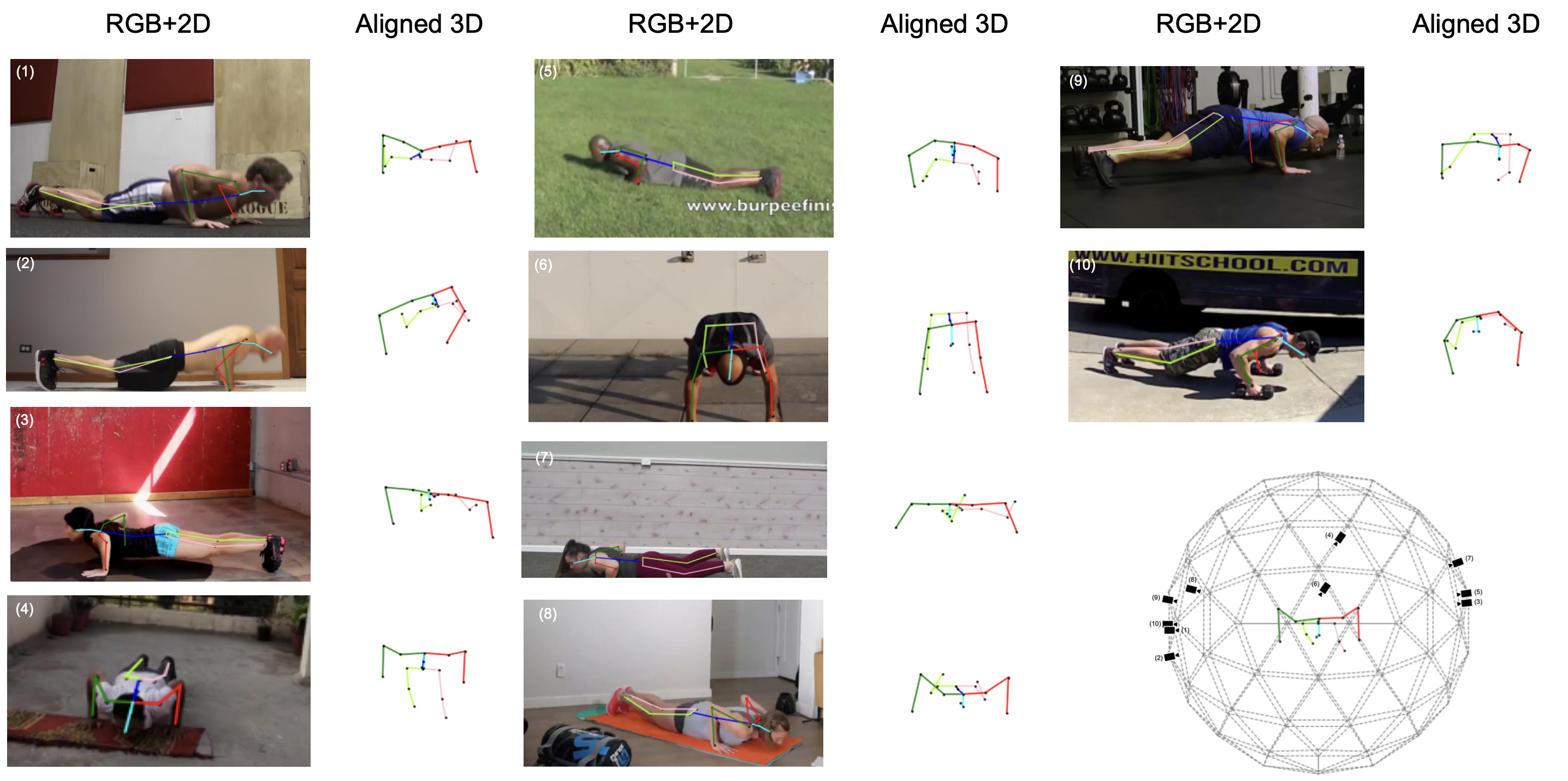}
  \caption{
  \textbf{2D vs 3D skeleton.} 3D skeleton is viewpoint invariant: all train/test 3D skeletons 
  in HAA4D are registered by their respective orthographic camera pose estimated using our model. Thus a single training 3D skeleton per class is sufficient, in stark contrast to the use of 2D RGB images or skeletons, where a large number of training data is required and uneven sampling is an issue. Trained with as few as one 3D+T or 4D {\em burpee} skeleton, the network can readily recognize {\em any} normalized burpee action shot from any viewing angles, including those under the person which are difficult to capture.
  }
  \label{fig:global_alignment_model}
  \vspace{-0.2in}
\end{figure}

\section{Introduction}


The goal of human action recognition in computer vision is to classify the person's action performed in a given video. Various approaches have been studied, such as RGB-based approaches~\cite{7843019, Cao_2020_CVPR} which utilize color images 
and recently use CNN-based or LSTM-based methods for classification, or 2D+T approaches~\cite{article} that encode 2D skeletons into some latent space for traditional supervised learning. 
The use of 3D skeletons provide outstanding results compared to other methods, as 3D skeletons are photometric, geometric and viewpoint invariant, so that as few as one single 4D skeleton is sufficient to learn the pertinent action class (Figure~\ref{fig:global_alignment_model}). These invariances  enable better temporal correlation by tracking the position of each 3D joint over time where each bone length is largely preserved throughout the action,  
in contrast to their projected 2D counterparts on images which may be distorted. 
Such invariances also make it easier for a computer algorithm to discern even fine differences in human poses. 

Given the above advantages of 3D skeletons, previous human action dataset contributions~\cite{hu2017jointly, liu2020ntu, 7415989, shahroudy2016ntu, smaira2020short} have  provided valuable skeletal information for action recognition. However, these datasets suffer inherent disadvantages, such as lacking poses diversity or relatively low ground-truth accuracy. HAA4D is dedicated in complementing 
current human skeleton action datasets by providing a more comprehensive range of {\em atomic} actions in real-life scenarios, which have demonstrated excellent performance in recognizing composite and complex actions~\cite{haa500}, while are easier to accurately annotate as each atomic action lasts for no more than 1--3 seconds. The samples in our dataset are collected from videos in-the-wild; therefore, we employ a deep alignment model to predict the camera pose in each video frame, followed by aligning all 3D+T skeletons in an uniform camera coordinate system (Figure~\ref{fig:global_alignment_model}) 
such that all faces are facing in the negative $z$-direction. As human actions in the same class often share similar trajectories in their skeletal movements, such global alignment which transforms all train/test examples to the same coordinate system can significantly improve matching the pertinent action sequences using explicit geometry features. 

With the emergence of deep learning and its superior performance in image classification tasks, related works mainly focus on supervised learning techniques that directly feed 4D human skeleton data into the network. Their networks are expected to generalize human actions and learn the underlying relationships in the embedded space. However, it is very difficult for deep learning to obtain satisfactory results with excessive learnable parameters when only a few examples are presented, given that 4D human skeletons are expensive to acquire. 
Furthermore, if the network is to recognize as many different human actions as possible, the size of the training dataset for training will be prohibitively large. 

\begin{figure}[t]
  \centering
  \hspace{-0.18in}
  \includegraphics[scale=0.24]{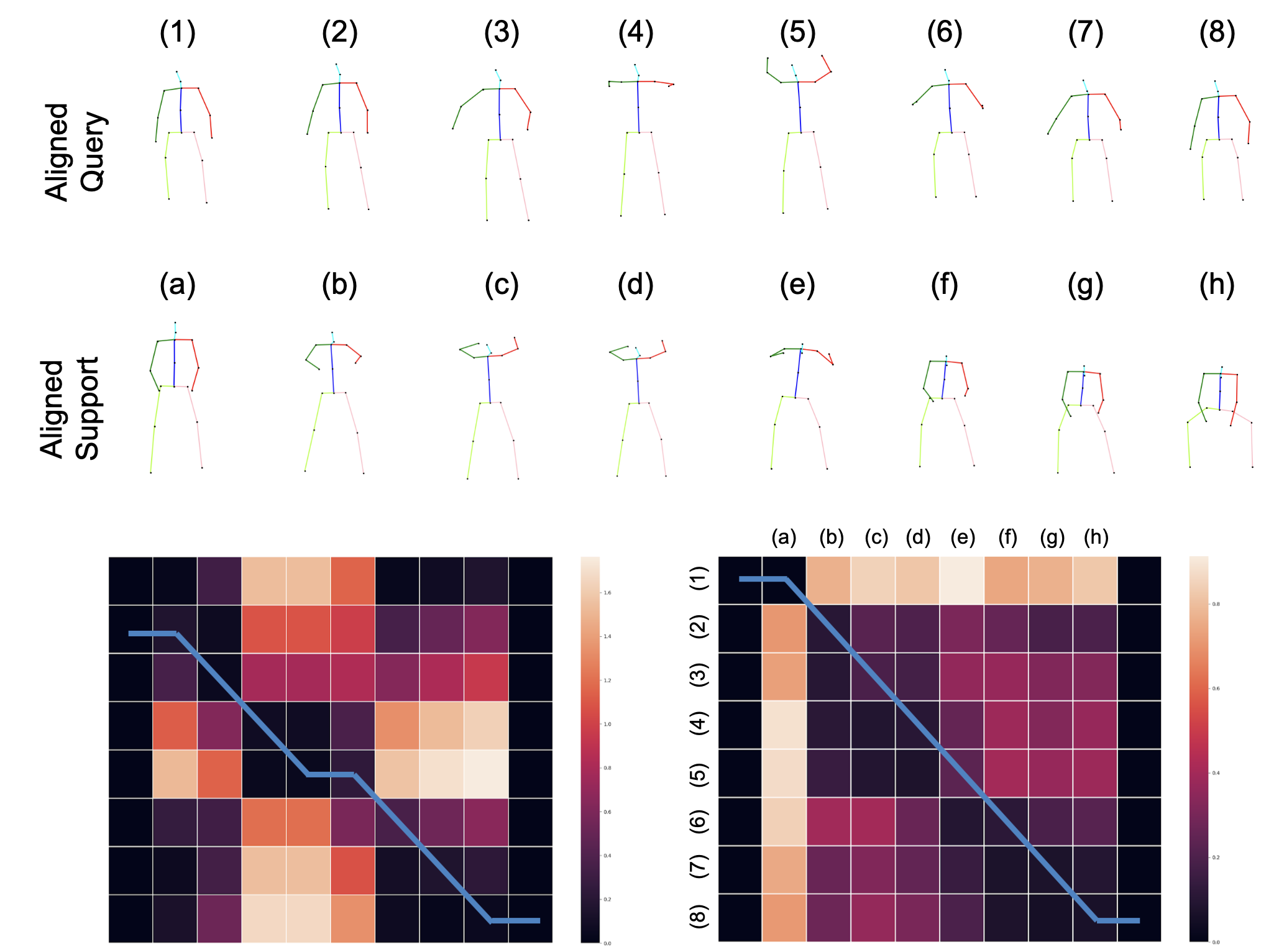}
  \caption{{\bf Implicit alignment} (left) {\bf vs explicit  alignment} (right). Alignment matrix of the  {\em battle-rope-power-slam} action between the query and  support videos. 
  The embeddings are fed to an OTAM module to generate a distance map of size $T\times T$ using the cosine distance between the query and supports.
  Darker cells indicate higher similarity, 
  while the blue line shows the matching path.}
  \label{fig:sample_skeletons}
  \vspace{-0.2in}
\end{figure}

Thus, our recognition model in this paper adopts {\em few-shot} learning techniques~\cite{bart2005cross, tarn, fei2006one, li2002rapid, miller2000learning}, making it the {\em first} on few-shot human atomic action recognition. Our few-shot model, combined with a {\em global alignment} model, performs matching using explicit 4D skeletal representation. Exploiting geometry inherent in 4D human skeletons avoids training complex models to learn proper encoding in the embedded space, which may still not produce better alignment than using explicit alignment (Figure~\ref{fig:sample_skeletons}) and thus hampering recognition. Our experiments validate that the proposed few-shot model working in tandem with a global alignment model 
produces better recognition results with fewer training examples for each action.
In summary, our contributions consist of:
\begin{enumerate}

\item HAA4D, a human atomic action dataset where all 4D skeletons are globally aligned. HAA4D complements 
existing prominent 3D+T human action datasets such as  NTU-RGB+D \cite{liu2020ntu, shahroudy2016ntu}, and Kinetics-skeleton \cite{kay2017kinetics};
HAA4D contains 3390 samples of human actions in the wild with 300 different kinds of activities. The samples for each human action range from 2 to 20, each provided with RGB frames and their corresponding 3D skeletons;

\item introducing an alignment network for predicting orthographic camera poses in the train/test samples, where all 4D skeletons are aligned in the same camera space each facing  the negative $z$-direction. This allows for better recognition results with small number of training samples compared to ST-GCN\cite{stgcn}, Shift-GCN\cite{shiftgcn}, and SGN\cite{zhang2020semantics};
\item the first few-shot baseline for 4D human (atomic) action recognition that produces results comparable to or better than state-of-the-art techniques on unseen classes using a small number of training samples.
\end{enumerate}

\begin{table*}[t]
    \begin{center}
    \resizebox{0.75\textwidth}{!}{%
    \begin{tabular}{l c c c c c } 
     \toprule
     Datasets & Samples & Actions & Views & Modalities & Year \\ [0.5ex] 
     \midrule
     UWA3D Multiview II~\cite{7415989} & 1075 & 30 & 5 & RGB+D+3DJoints & 2015 \\ 
     \hline 
     NTU RGB+D~\cite{shahroudy2016ntu} & 56,880 & 60 & 80 & RGB+D+IR+3DJoints & 2016 \\ 
     \hline
     SYSU 3DHOI~\cite{hu2017jointly} & 480 & 12 & 1 & RGB+D+3DJoints & 2017 \\
     \hline 
     Kinetics 400 ~\cite{kay2017kinetics} & 306,245 & 400 & - & RGB & 2017 \\
     \hline
     NTU RGB+D 120~\cite{liu2020ntu} & 114,480 & 120 & 155 & RGB+D+IR+3DJoints & 2019 \\
     \hline 
     Kinetics-700-2020 ~\cite{smaira2020short} & 633,728 & 700 & - & RGB & 2020 \\
     \hline 
     Ego4D~\cite{grauman2021ego4d} & 3,025 & 110 & - & RGB+3DMeshes+Audio & 2021 \\
     \hline
     HAA4D & 3,390 & 300 & 2/20 & RGB+3DJoints & 2021\\ 
     \bottomrule
    \end{tabular}%
    }
    \end{center}
      \caption{Comparison between  HAA4D and other public datasets for human action recognition. With fewer samples than many existing datasets, HAA4D provides class-balanced and diversified actions with 3D+T or 4D skeleton data useful for few-shot action learning.}
  \label{tab:main}
  \vspace{-0.2in}
\end{table*}
\section{Related Works}

\subsection{Human Video Action Datasets}

In skeletal human action recognition, several datasets are often used as benchmarks, see Table \ref{tab:main}.  NTU-RGB+D~\cite{shahroudy2016ntu} and NTU-RGB+D 120~\cite{liu2020ntu}  were collected under laboratory settings that cover actions in three main categories: daily actions, mutual actions, and medical conditions. These datasets comprise of RGB videos, depth map sequences, 3D skeletal data, and infrared videos for each sample. There are other RGB+D datasets that were also captured in laboratory settings, such as  UWA3D Multiview II~\cite{7415989} which contains 30 various activities, and SYSU 3D HOI~\cite{hu2017jointly} which provides 12 activities with 40 different participants focusing more on human-object interaction.   Recently,  Ego4D~\cite{grauman2021ego4d} has been released which contains massive-scale egocentric videos and  provides long sequences of actions  in real-life scenarios, where parts are given with the corresponding 3D meshes. However, although having a massive number of samples per action, these datasets are often not sufficiently diversified, subject to the drawback of not covering a wide range of different actions.

Kinetics-skeleton, on the other hand, contains more diversified human actions. This skeleton dataset was built on top of Kinetics \cite{kay2017kinetics}, a video action datasets collected from YouTube with more than 300,000 video clips covering 400 action classes. OpenPose~\cite{cao2019openpose} was used to extract 2D human skeletons. Together with EvoSkeleton~\cite{Li_2020_CVPR} used to lift the 2D key-points to 3D skeletons, we can extract the 3D skeletal data from the images. While Kinetics-skeleton introduces more variety of human poses to the human skeletal action recognition domain, the correctness of the 3D skeletons generated may be questionable, as the 2D joints prediction network can fail if parts of the human body are outside of the image frame. 

To ameliorate the problem of existing datasets problem, namely, lack of pose diversity and low ground-truth accuracy, we introduce HAA4D which contains 300 human action classes, each with 20 examples with accurately annotated 2D joints position. 

\subsection{Video-Based and Few-Shot Action Recognition}
For more recent examples on video-based action recognition, C3D~\cite{tran2015learning} performs 3-dimensional convolution on the input image sequence to extract spatio-temporal features. In~\cite{Ullah2018ActionRI} an architecture is proposed for first processing each image using 2D convolution, and then using a bidirectional LSTM network to learn temporal information. TSN~\cite{wang2016temporal}  divides a video into several segments and selects snippets to pass to a spatial stream ConvNets and a temporal stream ConvNets. Video-level prediction is then derived from the consensus of the snippets. However, all these models contain an excessive number of learnable parameters that require training on large-scale datasets, which can fail when training samples are few and expensive to obtain. Thus, more works have focused on few-shot video action classification.

In~\cite{mishra2018generative} the authors proposed to learn to generate class embeddings using a Word2Vec~\cite{word2vec} model. NGM~\cite{Guo_2018_ECCV} introduces a graph matching metric on a graphical representation of a video. Dense dilated network~\cite{xu2018dense} uses a dilated CNN network on videos. In contrast to these works that neglect the temporal dimension of videos, TARN~\cite{tarn} introduces temporal attention for temporal alignment of videos.  A similar approach is used in~\cite{perminv} by adopting temporal attention as a temporal form of self-supervision. Our work is closest in spirit to OTAM~\cite{otam} which uses an ordered temporal alignment module, inspired from dynamic time warping (DTW)~\cite{dtw}. However, instead of focusing only on temporal alignment, HAA4D explicitly aligns along 3D+T or 4D dimensions since similar actions, after calibration removing the view and scale variation, share similar joint distribution and moving sequences. 

However, all of the aforementioned approaches directly train a video classification model using RGB-frames, which poses a deep neural network a great challenge in relating video frames 
as they can vary widely in background and illumination. In this paper, we use skeleton data extracted from the videos for better spatio-temporal alignment.

\subsection{Skeleton-Based Action Recognition}

Given the aforementioned advantages, 
skeleton-based action recognition has been a popular topic in computer vision that aims to classify actions using skeleton joint information. Conventional methods rely on hand-crafted features to capture dynamic movement of human joints~\cite{hussein2013human,vemulapalli2014human, wang2012mining}. With emergence of deep-learning, earlier related works 
represent joint coordinates as a vector, and  use recurrent neural network for action classification~\cite{du2015hierarchical,liu2016spatio,zhang2017view}, or represent skeleton sequence as a pseudo-image as input to a CNN-based network~\cite{ke2017new,kim2017interpretable,li2017skeleton}.
Recently, ST-GCN in~\cite{stgcn}, which consists of spatial graph convolution layers and temporal convolution layers, uses graph representation of skeleton sequence. ST-GCN has been extended in~\cite{shiftgcn} by incorporating  shift convolution~\cite{jeon2018constructing,wu2018shift,zhong2018shift} on a graph CNN, which achieves better performance with a lighter network. In this paper, we use Shift-GCN~\cite{shiftgcn} as a backbone for our baseline model to compare the efficacy of our globally aligned 4D skeletons.

\section{HAA4D}
\begin{figure}[t]
  \hspace{-0.17in}
  \includegraphics[scale=0.19]{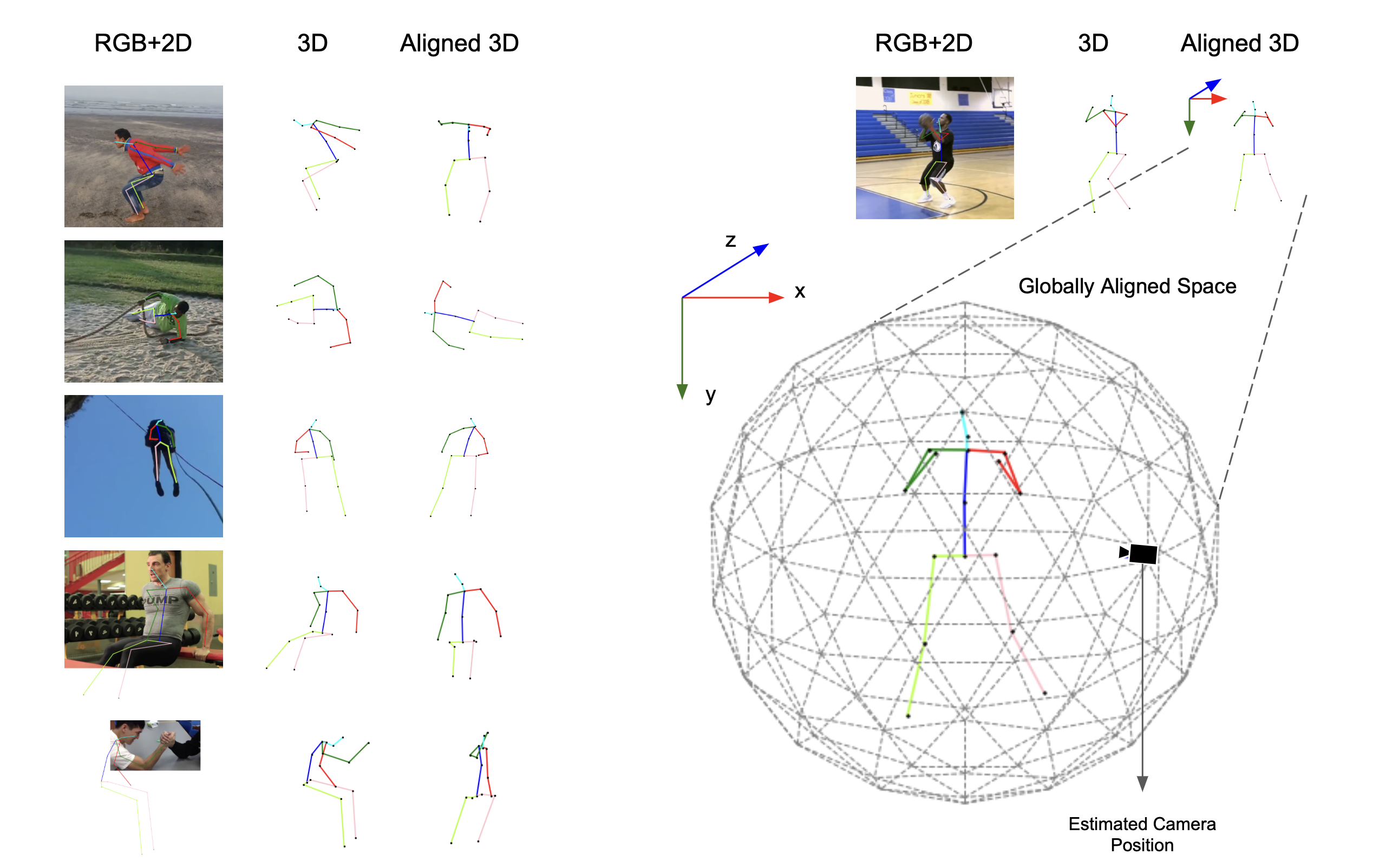}
  \caption{Samples from the HAA4D dataset. First column shows the RGB images and their 2D annotations, second and third columns represent the 3D skeletons in the original and globally aligned space (with face at negative $z$-direction). The right shows the predicted camera position at one of the subdivided icosahedron vertex in the globally aligned space.}
  \label{fig:skelelton-examples}
\end{figure}

\begin{figure*}[t]
  \centering
  \includegraphics[scale=0.3]{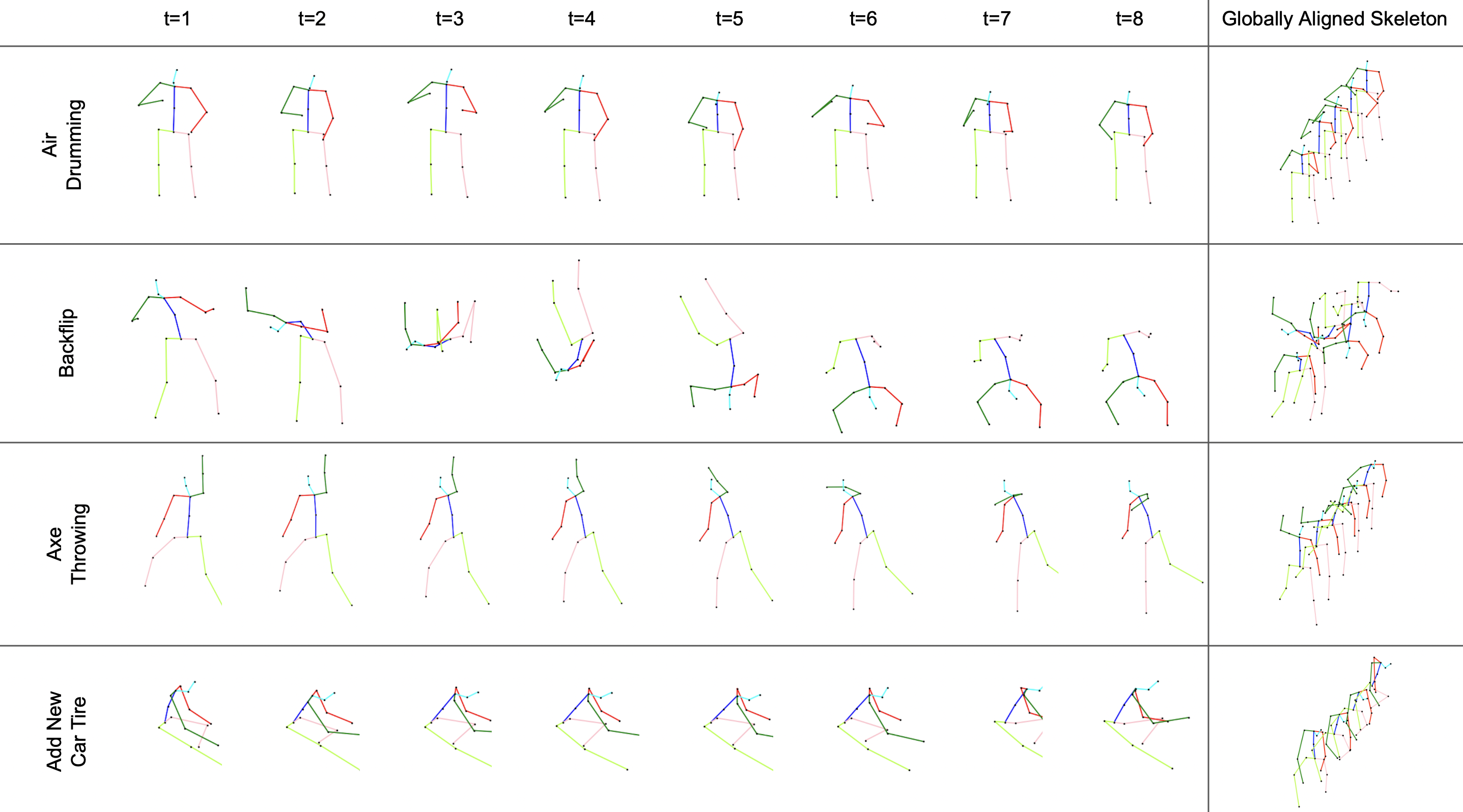}
  \caption{Examples of the 8-frame sampling of the HAA4D dataset.
  }
  \label{fig:eigth_frames_sampling}
  \vspace{-0.2in}
\end{figure*}

This section will introduce details and structure of our HAA4D dataset, its expandability, and the evaluation criteria. Figure \ref{fig:skelelton-examples} and \ref{fig:eigth_frames_sampling} shows some examples of our HAA4D dataset. Full details of HAA4D are available in the supplementary materials.

\begin{table}
  \centering 
  \resizebox{\columnwidth}{!}{%
  \begin{tabular}{lccc}
  
    \toprule
    Classes & Total Samples &  Globally Aligned Samples &  \\
    \hline
    300 & 3390 & 400 &\\ 
    \toprule
    Total Frames & Min/Max Frames & Average Frames &  \\
    \hline
    212042 & 7 / 757 & 63 & \\
    \toprule
    2 Examples & 20 Examples & Person Per Example & \\
    \hline
    145 classes & 155 classes & 1 &\\
    \bottomrule
    \end{tabular}%
    }
  \caption{Summary of HAA4D.}
  \label{tab:summary}
  \vspace{-0.2in}
\end{table}

\subsection{Dataset Structure}
HAA4D is a challenging human action recognition 3D+T dataset that is built on top of the HAA500 \cite{haa500} dataset. HAA500 is a curated video dataset consisting of atomic human action video clips, which provides  diversified poses with variation and examples closer to real-life activities. Refer to~\cite{haa500} for details on vocabulary selection and video clip selection, and advantages including clean labels for each frame, human-centric atomic-action frames, and high scalability. Table~\ref{tab:summary} tabulates a summary of HAA4D. 

Similar to Kinetics \cite{kay2017kinetics}, the original dataset provides solely in-the-wild RGB videos. However, instead of using existing 2D joints prediction networks, which often produce inaccurate results  when joints are hidden or not present, HAA4D consists of hand-labeled 2D human skeletons position. These accurate features can not only increase the precision of the 3D ground-truth skeleton but also benefit training for new joints prediction models that can reasonably hallucinate out-of-frame joints. The labeled 2D joints will then be raised to 3D using EvoSkeleton~\cite{Li_2020_CVPR}.
HAA4D is thus named with its accurate per-frame 3D spatial and temporal skeletons for human atomic actions. 

HAA4D consists of 3,300 real-life RGB videos of specific human action classes with the corresponding 2D joints and 3D human skeletons. The skeleton topology follows the one in Evoskeleton~\cite{Li_2020_CVPR} which consists of 17 joints which are quite sufficient in representing a wide range of human body and part movements. The dataset has 300 action classes that are divided into two categories: primary classes and additional classes. Primary classes have 20 samples per class, while additional classes contain two examples per class. Recall the main advantage of 4D skeleton data is its viewpoint invariance with a few as one 3D skeleton per frame, given they are globally aligned with camera poses available which are estimated using our camera prediction model.

For primary classes, videos 0--9 are used for training, and the remaining are for evaluation. Actions in the additional classes are dedicated to one-shot learning that evaluates the performance of the proposed model to differentiate the action when very limited data are presented. For example, this paper trains the alignment model on the first ten examples of human movements in the primary classes with data augmentation techniques to be detailed. The alignment model is then tested with videos in the evaluation sets of the primary classes for cross-view evaluation, and with examples in the additional classes for cross-action assessment. 

\begin{figure}[b]
  \centering
  \includegraphics[scale=0.18]{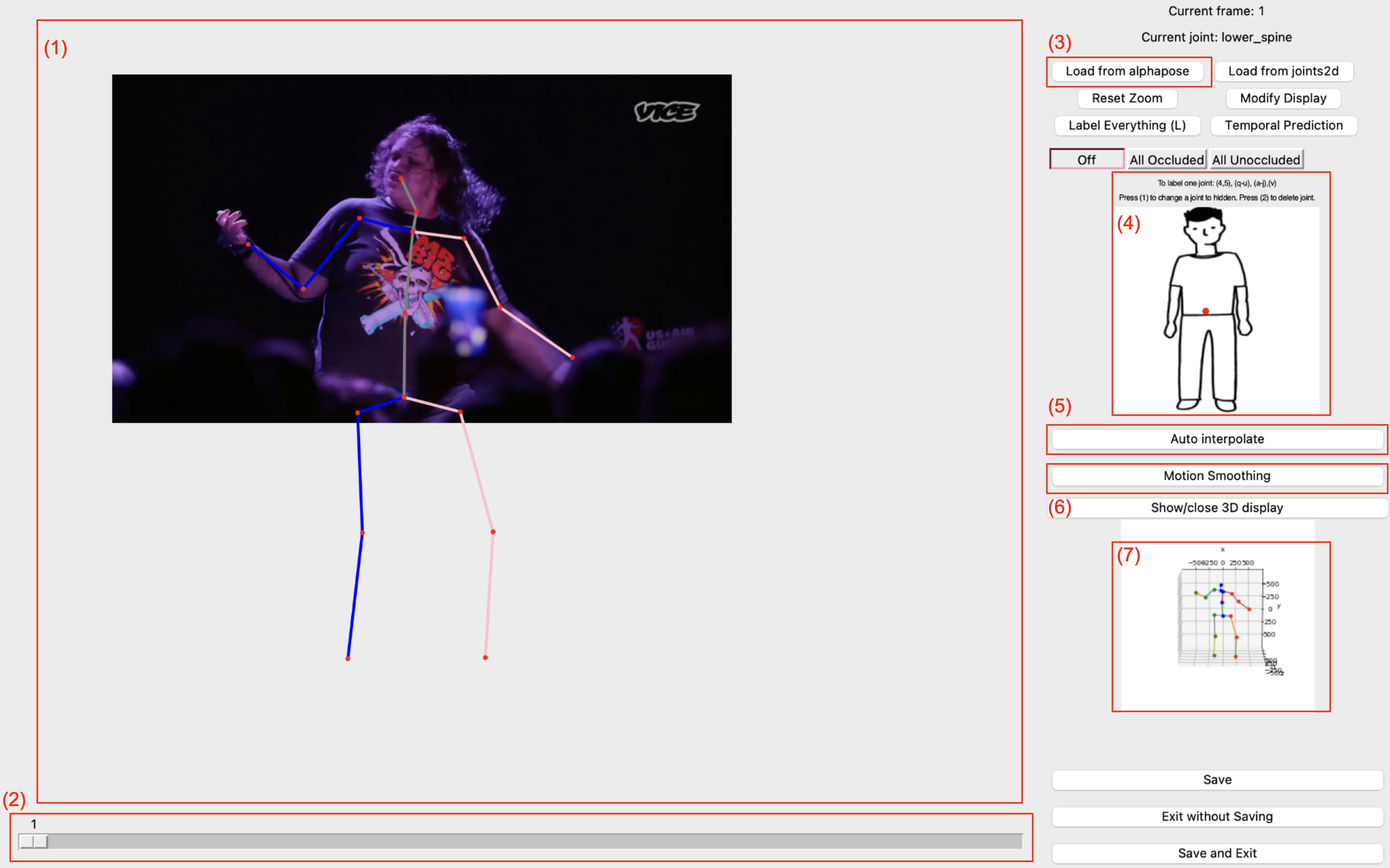}
  \caption{Main functionalities of our HAA4D annotation tool. (1): Annotating Canvas. (2): Dragging bar to skim through all the frames. (3): Load preliminary prediction from AlphaPose. (4): Map of the current label joint. (5): Linear interpolation. (6): Gaussian smooth the trajectories of all the joints. (7): Predicted 3D position from EvoSkeleton.}
  \label{fig:annotation}
\end{figure}

\subsection{Mean Average Precision (mAP)}
We calculated the mean average precision(mAP) of the whole dataset and each primary class using the bounding boxes of the predicted skeletons generated from AlphaPose~\cite{fang2017rmpe} and our human-labeled ground truth skeletons to calculate the IoU. The threshold of the IoU is set to 0.5. The overall mAP of the HAA4D dataset is 63.71. However, we observed a significant variance between the mAP of each class. For example, the mAP of the action `abseiling` is 92.08 while the mAP of `applying\_cream` is 1.28. One reason that caused the difference is that most images captured contain only the upper half of the body for activities like `applying\_cream`. This shows the insufficiency of state-of-the-art pose prediction networks like AlphaPose in predicting joints that are out of boundary; therefore, making our perfectly labeled ground truth more valuable. Although bounding boxes cannot effectively represent each joint's position accuracy, we will provide part-level bounding annotation for more accurate evaluation. The mAP details for each action class will be included in the supplemental material.

\subsection{Annotation Tool}
To help expand the dataset to more actions and classes, we have developed a simple interactive annotation tool that supports faster user labeling (Figure~\ref{fig:annotation}). Similar to Kinetics-skeleton use of  OpenPose~\cite{cao2019openpose} to predict positions of 2D joints, the user can load preliminary predictions from AlphaPose~\cite{fang2017rmpe} and can  correct intermediate frames where the network prediction is not precise. Observing that human movement when viewed in a short period is relatively linear, the annotation tool supports linear interpolation so that the user can avoid frame-by-frame processing whenever possible. 
The predicted 3D skeleton is also shown alongside in the interface for the user to monitor the generated ground truth in real time. The annotation tool covers the end-to-end generation of 3D human skeletons from images, which is user-friendly and efficient in annotation, thus making this HAA4D dataset easily expandable. We hope that by providing this annotation tool, the dataset can become more comprehensive in the future to  cover more human action poses, contributing to research in few-shot human pose estimation and action recognition that utilizes 3D+T data.

\begin{figure*}[t]
  \centering
  \includegraphics[scale=0.33]{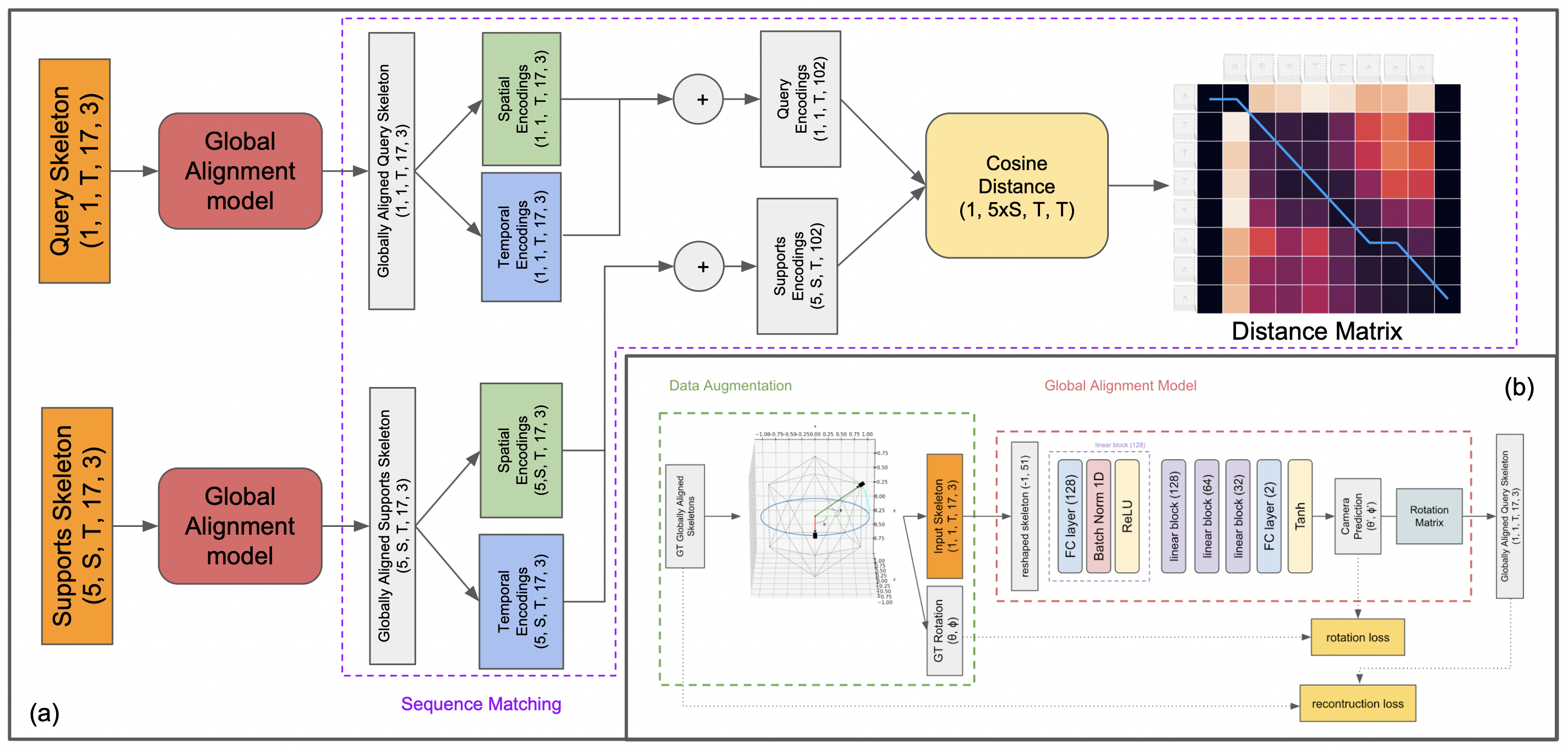}
  \caption{(a) Overview of the globally aligned few-shot action recognition architecture. (b) Architecture of the Global Alignment Model together with data augmentation.}
  \label{fig:our-model}
  \vspace{-0.2in}
\end{figure*}

\subsection{Globally Aligned Skeletons}
\label{sec:3.3}
Since examples in HAA500 were shot in the wild, each action sample has different viewpoints and scales, making the predicted 3D skeletons in various coordinate systems. Therefore, in HAA4D, we apply the following to standardize the 3D generated skeletons. 

First, all skeletons are centered such that their lower-spine joint is placed at the origin.  Second, because EvoSkeleton~\cite{Li_2020_CVPR} generates 3D skeletons frame-by-frame, it does not guarantee equal bone lengths throughout the video. Hence, to ensure the human skeleton's uniformity in the action video, we use bone lengths in the first frame as the reference to replace the lengths of the rest frames. We also scale all the bones such that their l2-norm equals to one, i.e., $b^f_i$ represents bone $i$ in the $f$-th skeleton frame and $B$ is the set of bones in the skeleton topology.

Then, we provide globally aligned skeletons (to parts of the dataset) by manually rotating them so that the human faces the negative $z$-direction at the start of each action. We choose the action to face in the negative $z$-direction because most people when taking photos will look into the camera. We calculate the alignment using only the first frame and apply the rotation to the rest of the sequence as a frame-by-frame alignment will break the relative movement of the original action. The calibration makes our data view-invariant by bringing all skeletons to the same space. We will argue in the next section that eliminating the rotation and scale factors help us better compare the skeleton movement in different actions by directly using the adjusted 3D joints position. The manually corrected, globally aligned skeletons are also used to train the alignment model, which takes any actions, regardless of their view angles and action type, and transforms them to the global coordinate system.

\section{Few-Shot Action Recognition Model}

In this section, we describe our few-shot skeleton-based action recognition model. Figure~\ref{fig:our-model} shows the overall architecture which consists of two modules: the global alignment module (with data augmentation) and the sequence matching module. 

Following~\cite{otam,wang2016temporal}, given a video example, we divide it into $t_n$ segments and randomly select a frame within each segment. We randomly selected one frame from each segment to introduce more randomness to each input sequence. As we work on atomic actions, given the same action type, the start and end will match. However, each person may act at a different pace, and thus, we do not hardcode any frame to represent each segment. 

We then concatenate the $t_n$ skeletons to represent the video, denoted by $S$. This allows each video to be represented by the same number of frames. Similarly, the support sets are constructed by randomly selecting $s_n$ samples from each $w_n$ class, including the query's. The global alignment model transforms both the query and support into the globally aligned space before  action recognition, by comparing the sequence matching results with the support sets.   

\subsection{Global Alignment Model (GAM)}
\label{sec:4.1}
As query and support videos can be captured in-the-wild, we propose the global alignment model to bring all inputs to the same space, such that actions face in the negative $z$-direction at the beginning. The camera angle will also be available after our global alignment. 

To prepare or synthesize training data, since we have the ground truth globally aligned skeletons as described in section~\ref{sec:3.3}, we perform data augmentation by uniformly sampling different camera views from vertices on an $n$-frequency icosahedron (i.e., subdivided icosahedron incident on a unit sphere as shown in Figure~\ref{fig:global_alignment_model}).
Our globally aligned skeletons are viewed from the default orthographic camera $C_D$ located at $(0,0,-1)$. By simplifying the camera  to have no self-rotation, we only need to predict two rotation parameters, namely, azimuth and altitude angles, denoted by $\beta$ ($\theta$ and $\phi$) with respect to the default camera position $(0,0,-1)$.
We obtain $\theta$ and $\phi$ by:

\begin{gather}
\theta = \pi - \arctan\left({\frac{C_x}{C_z}}\right),\quad \theta = \begin{cases}
    \theta - 2\pi,& \text{if } \theta > \pi\\
    \theta, & \text{otherwise}
\end{cases} \\
\phi = \arctan\left({\frac{C_y}{\sqrt{(C_x)^2+(C_z)^2}}}\right)
\end{gather}

We then use  Rodrigues rotation to obtain the rotation matrix $R=R_2 \cdot R_1$, where:
\begin{equation}
\begin{split}
\text{Rodrigues}(\hat{n}, \alpha) = I_3 + \sin{\alpha}K + (1-\cos{\alpha})K^2, \\
\textrm{where }  K = \text{Skew}(\hat{n})
\end{split}
\end{equation}

We first rotate in $\theta$ using:
\begin{equation}
R_1 = \text{Rodrigues}(\hat{n}_1, \theta), \quad \hat{n}_1 = \left\langle
\begin{matrix}
0 & -1 & 0
\end{matrix}
\right\rangle
\end{equation}
Then rotate in $\phi$: let
$\vec{v}_1 = \left\langle
\begin{matrix}
0 & 1 & 0
\end{matrix}
\right\rangle, \quad \vec{v}_2 = R_1 \cdot C_{D}
$,
then
\begin{equation}
\begin{split}
R_2 = \text{Rodrigues}(\hat{n}_2, |\phi|), \textrm{where} \\
\hat{n}_2 = \begin{cases}
    \vec{v}_2 \times \vec{v}_1,& \text{if } \phi > 0\\
    \vec{v}_1 \times \vec{v}_2, & \text{otherwise}
\end{cases}
\end{split}
\end{equation}

We obtain the augmented view skeleton $S_{A}$ by rotating the globally aligned skeletons $S_{G}$:
\begin{equation}
S_{A} = R^{-1} \cdot S_{G} \textrm{, where } R = R_2 \cdot R_1
\end{equation}

We sample data from 92 camera angles on a 3-frequency subdivision icosahedron, trained on 73 views, and test on the rest 19 views. We employ rotation loss and reconstruction loss defined as follows:
\begin{eqnarray}
 \mathcal{L}_{rot} = \text{MSE}(\cos{\beta_{pr}}, \cos{\beta_{gt}}) 
 + \text{MSE}(\sin{\beta_{pr}}, \sin{\beta_{gt}})\\
 \mathcal{L}_{rec} = \text{RMSE}(S_{pr},S_{A}), \textrm{where~} 
 S_{pr} = (R(\beta_{pr}))^{-1} \cdot S_{G}
\end{eqnarray}

The final loss for our global alignment model is
\begin{equation}
\mathcal{L} = w_1\mathcal{L}_{rot} + w_2\mathcal{L}_{rec},
\end{equation}
where we set $w_1$ to 10 and $w_2$ to 1.

\subsection{Sequence Matching}
After transforming all input skeletons to the same global space, where we denote each transformed skeleton as $S$, we encode the 3D positions of the joints $f_{\mathit{pos}}(S)$ and  their respective trajectory $f_{\mathit{tra}}(S)$ as embeddings for sequence matching. The encoding functions are defined as follows:

\begin{gather}
f_{\mathit{pos}}(S) = S^{1} \oplus S^{2} \oplus \cdots \oplus S^{t_n} \\
f_{\mathit{tra}}(S) = \mathds{1} \oplus (S^{2}-S^{1}) \oplus \cdots \oplus (S^{t_n} - S^{t_{n-1}}) \\
f_{\mathit{skel}}(S) = f_{\mathit{pos}}(S) \oplus f_{\mathit{tra}}(S)
\end{gather}
where $\oplus$ denotes concatenation between the two frames.
Inspired by OTAM~\cite{otam} and dynamic time warping~\cite{dtw}, we use the matching score between the explicit query $f_{\mathit{skel}}(S_{q})$ and supports $f_{\mathit{skel}}(S_{s})$ in embedding to determine the action class of the given query. We generate the distance matrix $D = 1 - \text{cosine\_similarity}(f_{\mathit{skel}}(S_{q}), f_{\mathit{skel}}(S_{s}))$. The score of the query and the support is represented by $-\sum_{d\in \mathit{path}}d$, where $\mathit{path}$ is a relaxed-minimum path of $D$ defined in~\cite{otam}. The final score of a query  and a support class is the average of scores for each $k$ support data of the class. The top final score corresponds to the action class of the query video.

\section{Experiments}

\subsection{Skeletal Representation}

\paragraph{Implicit vs Explicit}
Without representative skeletal-based few-shot action recognition models, we create a baseline model to evaluate the performance. The baseline model uses Shift-GCN~\cite{shiftgcn} as the backbone to encode input skeletons in an embedded space and perform the temporal alignment. The Shift-GCN model is first pre-trained with the embedding cosine similarity loss and the skeleton reconstruction loss. To simulate the few sample situations in the dataset, we randomly select ten samples from each NTU-RGB+D~\cite{shahroudy2016ntu} and for training, and use the other ten examples for testing. The encoded skeleton in the embedded space is the implicit representation of the skeleton sequence used for sequence matching.

\vspace{-0.15in}
\paragraph{2D vs 3D}
Here we compare the effectiveness between the 2D and 3D skeletons. The most notable difference between using 2D and 3D skeletons is that for 2D skeletons, we cannot perform data augmentation as rotation cannot be done in the 2D space without breaking the original skeleton structure. We experiment on both the baseline model and other proposed architectures. In the baseline model, we follow the same approach to encode the 2D skeletons to an implicit representation in the embedded space. However, for testing on our model, we directly feed 2D skeletons for sequence matching without the proposed 3D global alignment. 

\vspace{-0.15in}
\paragraph{Results}
Table \ref{tab:comparison-baseline} shows the effectiveness of using different representations of human skeleton sequences for action recognition. The baseline model utilizes implicit representation, while our model uses explicit information. We test both models with 2D and 3D as input skeletons and demonstrate that explicit 3D representation, i.e., globally aligned information, produces the best accuracies in action recognition. 

\begin{table}
  \begin{center}
  \resizebox{\columnwidth}{!}{%
  \begin{tabular}{lcccc}
    \toprule
    & \multicolumn{2}{c}{NTU-RGB+D xview \cite{shahroudy2016ntu}} & \multicolumn{2}{c}{HAA4D}     \\
                                  & 5way 1shot & 5way 5shot & 5way 1shot & 5way 5shot \\
    \midrule
    Baseline (I) \\
    \quad - 2D              &  24.7    &   25.5  &   23.4     &  25.6   \\
    \quad - 3D              &  46.8   &   51.7  &  47.0     & 50.3 \\
    \cline{1-5}
    Ours (E) \\
    \quad - 2D              &  44.7    &   49.8  &  45.8     &  48.3   \\
    \quad - 3D              & \bf{57.8}   &  \bf{70.0}  & \bf{52.1}     &  \bf{62.3}  \\
    \bottomrule
  \end{tabular}%
  }
  \end{center}
  \caption{Performance under different skeletal representation. (I) stands for using implicit representation, and (E) means using explicit representation of human skeleton.}
  \label{tab:comparison-baseline}
  \vspace{-0.2in}
\end{table}


\subsection{Ablation Studies}

In this section, we conduct ablation studies to examine the effect of individual part of our model. We tested on three different factors: the number of sampled frames, the sequence matching approach, and the explicit information used as encodings.  We monitor the results on both HAA4D and NTU-RGB+D.

\vspace{-0.15in}
\paragraph{Sampling Rate}

For each human skeleton action sequence, we divide them into several segments and select one skeleton from each to represent the segment. This sampling technique allows all the action sequences to be in uniform length while increasing the difference between the neighboring frames. We test on different numbers of frames to represent the sequence, as oversampling can lead to redundancy and unnecessary computational load, 
while undersampling can lead to  temporal aliasing. We padded the last frame to extend the input sequence for input with fewer frames than the segment size.

\vspace{-0.15in}
\paragraph{Sequence Matching}
Our action recognition model determines the action by comparing the sequence matching score between the query set and support sets. We experiment the following matching techniques:

Mean, Regular DTW, and OTAM~\cite{dtw, otam}. The Mean approach for  sequence matching  directly computes the average between the two distance matrix $D$ and thus neglects the temporal ordering. The other two methods consider  temporal ordering. Regular DTW assumes the two sequence matches at the start and end, which allows movement in the direction of $\searrow$, $\rightarrow$, and $\downarrow$ when computing the aligned path in the distance matrix. OTAM on the other hand assumes the alignment can start and end from any time in the sequence, thus only allowing $\searrow$ and $\rightarrow$ to ensure that all the alignment paths have equal length.

\begin{table}
  \begin{center}
  \resizebox{\columnwidth}{!}{%
  \begin{tabular}{lcccc}
    \toprule
    & \multicolumn{2}{c}{NTU-RGB+D xview \cite{shahroudy2016ntu}} & \multicolumn{2}{c}{HAA4D}     \\
                                  & 5way 1shot & 5way 5shot & 5way 1shot & 5way 5shot \\
    \midrule
    Mean              &      &     &            &     \\
    \quad - $t_n=8$ & 47.5      & 51.0        & 45.9 & 54.8    \\
    \quad - $t_n=32$ & 51.8      & 59.8         & 47.9 & 56.5   \\
    \quad - $t_n=\text{all}$ & 52.0      & 60.5        & 47.5 & 53.2    \\
    \cline{1-5}
    Regular DTW &       &      &             &      \\
    \quad - $t_n=8$ & 58.0      & 69.8       & 52.4 & 62.3   \\
    \quad - $t_n=32$ & \bf{59.3}      & \bf{71.3}       &  \bf{53.5} & \bf{62.7}    \\
    \quad - $t_n=\text{all}$ & 59.1      & 69.0       & 53.4 & 62.5    \\
    \cline{1-5}
    OTAM                                 &         &    &       &       \\
    \quad - $t_n=8$ & 56.5      &  67.8        & 51.9 & 62.0    \\
    \quad - $t_n=32$ & 57.8      & 70.0        & 52.1 & 62.3   \\
    \quad - $t_n=\text{all}$ & -      & -        & - & -    \\
    \bottomrule
  \end{tabular}%
  }
  \end{center}
  \caption{Performance of few-shot action classification under different sequence matching methods and sampling rates. We do not consider $t_n=\text{all}$ for OTAM as it constrains the distant matrix to be squared and having equal length paths.  }
  \label{tab:comparison-sequence-frames}
  \vspace{-0.1in}
\end{table}

\vspace{-0.15in}
\paragraph{Explicit Encodings}
In addition to directly using the aligned 3D coordinates ($0^{th}$ order) of the skeleton sequence as our explicit embeddings, we evaluate the effect of using different combinations between the $0^{th}$, $1^{st}$, and $2^{nd}$ order representations. The $1^{st}$ order of the skeleton sequence consists of the joints movement throughout temporal frames. The $2^{nd}$ order representation includes the curvature of the joint path. 

\vspace{-0.15in}
\paragraph{Results}
From our experimental results tabulated in Table~\ref{tab:comparison-sequence-frames}, the 32-frame sampling outperforms others in both HAA4D and NTU-RGB+D datasets. In addition, we observe that techniques that consider temporal ordering, such as regular DTW and OTAM, tend to produce better accuracy in action prediction for the sequence matching methods. Notably, the actions in HAA4D are atomic, which are diverse including atomic motions that form more complex actions, where the  actions in the same class match at the beginning as well as the end of the sequence.
This feature benefits the regular DTW to obtain more reliable results. 

Table~\ref{tab:comparison-explicit-encoding} shows that by using both the globally aligned position and the joints temporal movement, the model produces the best performance. 

\begin{table}
  \begin{center}
  \resizebox{0.9\columnwidth}{!}{%
  \begin{tabular}{lcccc}
    \toprule
    & $0^{th}$ only  & $0^{th}$+$1^{st}$ & $0^{th}$+$1^{st}$+$2^{nd}$   \\
    \midrule
    HAA4D \\
    \quad - 5way 1shot & 51.4 & \bf{52.1} & 50.7 \\
    \quad - 5way 5shot & 56.8 &\bf{62.3} & 59.6 \\
    \bottomrule
  \end{tabular}%
  }
  \end{center}
  \caption{Performance of the accuracy between using different explicit encoding.}
  \label{tab:comparison-explicit-encoding}
  \vspace{-0.2in}
\end{table}

\subsection{Comparison with  State-of-the-Arts}

This section evaluates state-of-the-art skeleton-based action recognition models on our HAA4D dataset, and compares the results of the testing on other public datasets. Although few-shot learning has been used on action recognition \cite{otam}, few approaches have been conducted via the 3D or 3D+T skeleton-based method. In addition, one of our claims is the use of few-shot learning on 3D skeleton human action recognition to improve the prediction accuracy in sparse training training data. We compare our result with the state-of-the-art supervised approaches to support our claim.

The skeletons of HAA4D are processed according to the description in section~\ref{sec:3.3}, which centers the origin and constrains all skeletons to have equal bone length throughout the video. The inputs of ST-GCN~\cite{stgcn} and Shift-GCN ~\cite{shiftgcn} follow the shape (batch-size $\times$ channels $\times$ frames $\times$ joints $\times$ people). We let the channels be the $(X, Y, Z)$ of coordinates of the 3D skeletons and pad zeros to the second person in people dimensions, as HAA4D targets only single-person action. SGN~\cite{zhang2020semantics} splits a multi-person frame into multiple frames, in which each frame contains only a single person. This model also requires segmenting the skeleton sequence into~20 clips, where one frame is randomly sampled from each segment to form the new sequence. We follow the same data processing techniques to evaluate our HAA4D.

Table~\ref{tab:comparison-state-of-the-art} shows that all models suffer from the lack of accuracy in Kinetics-skeleton. One possible reason is that the skeletons are directly extracted from the Kinetics using OpenPose~\cite{cao2019openpose} without correction done on the corrupted skeletons, where the 2D predictions are not sufficiently accurate to be lifted to the corresponding 3D skeleton. On the other hand, our HAA4D is a curated dataset with precise 2D joints location; even out-of-frame joints are manually completed reasonably. Thus the ground truth 3D skeleton is more reliable. This leads to better accuracy for the state-of-the-art models on our dataset. However, noting that HAA4D is designed for few-shot action recognition and due to the limited size of HAA4D, there is still a gap in prediction precision between HAA4D and NTU-RGB+D~\cite{shahroudy2016ntu}. 

\begin{table}
  \begin{center}
  \resizebox{\columnwidth}{!}{%
    \begin{tabular}{lccc}
    \toprule
    &   ST-GCN~\cite{stgcn} & Shift-GCN~\cite{shiftgcn} & SGN ~\cite{zhang2020semantics}\\
    \midrule
    NTU-RGB+D~\cite{shahroudy2016ntu} &&&      \\
    \quad- xview & 88.8 & {\bf 95.1} & 94.5 \\
    \quad- xsub & 81.6 & 87.8 & {\bf 89.0} \\
    \cline{1-4}
    Kinetics-skeleton~\cite{kay2017kinetics} & {\bf 31.6} & 17.7   & 20.8    \\
    \cline{1-4}
    Ours (HAA4D) & 21.2  & 53.0  & {\bf53.3}\\
    \bottomrule
  \end{tabular}%
  }
  \end{center}
  \caption{Performance of state-of-the-art skeleton-based action recognition models on different datasets.}
  \label{tab:comparison-state-of-the-art}
  \vspace{-0.2in}
\end{table}

\section{Discussion}
\subsection{Substantial need for the clean HAA4D dataset}
One may concerned that ``clean" dataset is not practical in real-life scenarios. However, current 2D pose prediction networks are still far from the accuracy of hand-corrected labeling. With much effort, introducing this cleaner dataset can advance this area of studies as all images collected in HAA4D are ``in the wild" with more accurate ground truth 2D skeletons. Moreover, we observed several actions in NTU RGB+D 120 \cite{liu2020ntu} and Kinetics-skeleton \cite{kay2017kinetics} are corrupted, including wrong subjects or unreasonable joint movements, that humans cannot even recognize. Our highly diversified HAA4D does not have these problems to distract learning. Also, having a clean dataset, we can add and control noises to generate false actions and even train networks to identify these errors.

\subsection{Evaluation on failure cases}
\begin{figure}
  \centering
  \includegraphics[scale=0.27]{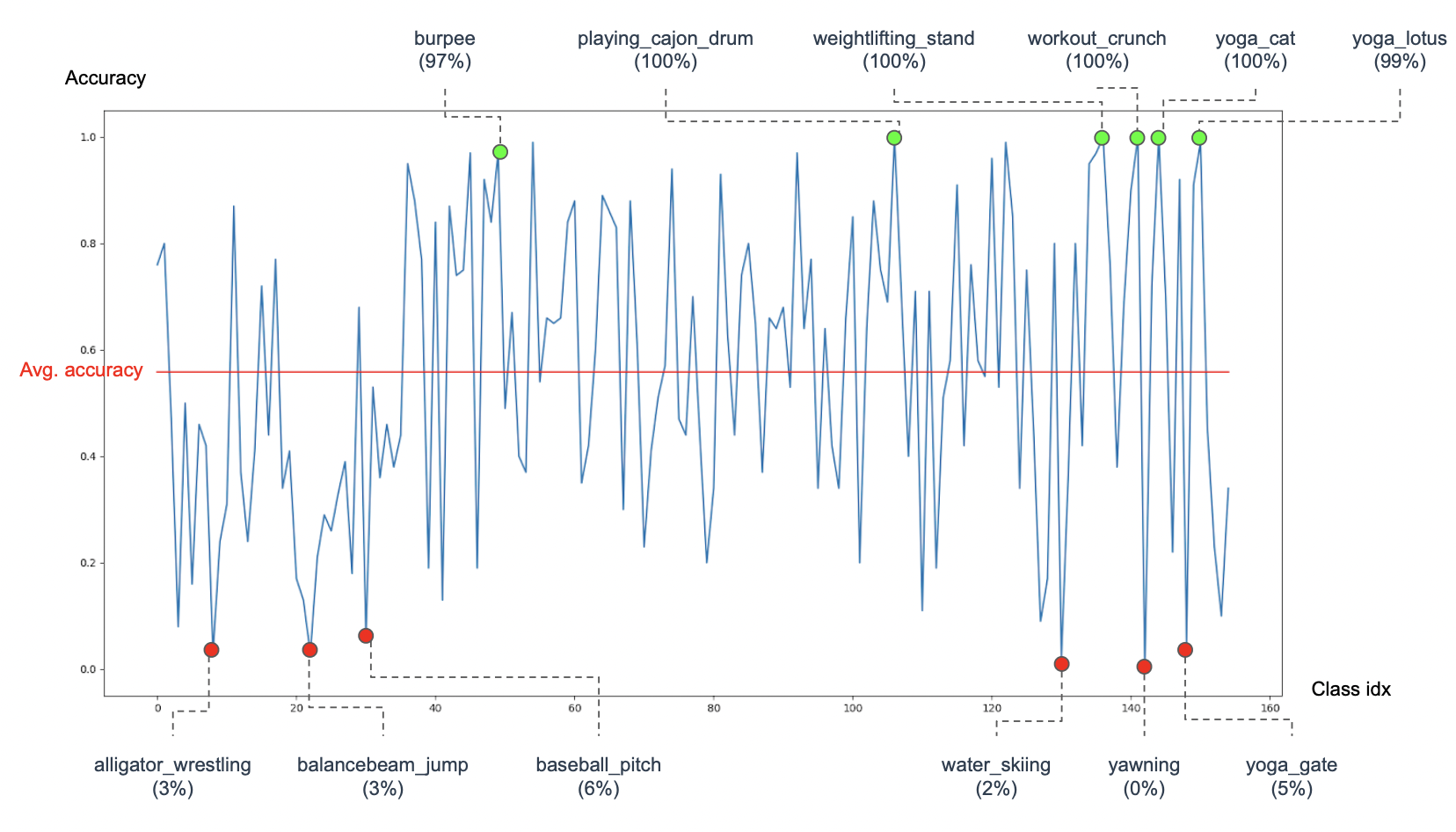}
  \caption{Accuracy respect to each action classes}
  \label{fig:in-depth analysis}
  \vspace{-0.2in}
\end{figure}

Figure~\ref{fig:in-depth analysis} shows the worst 5 classes for our action recognition model are yawning (0\%), water\_skiing (2\%), alligator\_wrestling (3\%), balancebeam\_jump (3\%), yoga\_gate (5\%). Actions such as yawning mainly involve mouth movements, which is difficult to reflect using skeletons. The model is subject to cases that lack significant movements in parts of the skeleton. Similar cases can be found in water\_skiing and alligator\_wrestling. The model also obtains low accuracy when the actions contain significant variances and lack consistency. For actions like balancebeam\_jump, different dancers  have unique styles and thus poses. It is the balance beam (interacting object) that helps humans identify the action even though each example looks completely different.

\section{Conclusion}

This paper proposes a new 4D dataset {\bf HAA4D} which consists of more than {\bf 3300} RGB videos in {\bf 300} human atomic action classes. HAA4D is clean, diverse, class-balanced where each class is viewpoint-balanced with the use of 3D+T or 4D skeletons. All training and testing 3D skeletons in HAA4D are globally aligned. Such alignment makes matching skeletons more stable, and thus with fewer training samples per class needed for action recognition leveraging full 3D+T information, in stark contrast to the 2D counterparts where massive amount is required to cover adequate viewpoints.  

Given the high diversity and skeletal alignment in HAA4D, we construct the first baseline few-shot 4D human atomic action recognition network, which produces comparable or higher performance than relevant state-of-the-art techniques, using the same small number of training samples of unseen classes. Through ablation studies, we have verified the advantages of 3D skeletons over 2D counterparts. We have also found that existing 3D skeletal action datasets (e.g., NTU-RGB+D \cite{shahroudy2016ntu} and Kinetics-skeleton \cite{kay2017kinetics}) significantly benefit from the proposed explicit alignment in the few-shot setting. With its high scalability where only a small number of 3D+T skeletons to be annotated for class extension using our annotation tool, and 3D+T global alignment  contributed by this paper, we hope HAA4D will spawn fruitful future works on skeletal human action recognition. Dataset, codes, models and tools will be released upon publication.

\section{Acknowledgment}
This research was supported by the Research Grant Council, Hong Kong SAR under grant no. 16201420.

{\small
\bibliographystyle{ieee_fullname}
\bibliography{cvpr}

\begin{thebibliography}{10}\itemsep=-1pt

\bibitem{dtw}
{\em Dynamic Time Warping}, pages 69--84.
\newblock Springer Berlin Heidelberg, Berlin, Heidelberg, 2007.

\bibitem{bart2005cross}
Evgeniy Bart and Shimon Ullman.
\newblock Cross-generalization: Learning novel classes from a single example by
  feature replacement.
\newblock In {\em CVPR 2005}.

\bibitem{7843019}
Amel Ben~Mahjoub and Mohamed Atri.
\newblock Human action recognition using rgb data.
\newblock In {\em 2016 11th International Design Test Symposium (IDT)}, pages
  83--87, 2016.

\bibitem{tarn}
Mina Bishay, Georgios Zoumpourlis, and Ioannis Patras.
\newblock Tarn: Temporal attentive relation network for few-shot and zero-shot
  action recognition.
\newblock In {\em BMVC 2019}.

\bibitem{otam}
Kaidi Cao, Jingwei Ji, Zhangjie Cao, Chien-Yi Chang, and Juan~Carlos Niebles.
\newblock Few-shot video classification via temporal alignment.
\newblock In {\em CVPR 2020}.

\bibitem{Cao_2020_CVPR}
Kaidi Cao, Jingwei Ji, Zhangjie Cao, Chien-Yi Chang, and Juan~Carlos Niebles.
\newblock Few-shot video classification via temporal alignment.
\newblock In {\em Proceedings of the IEEE/CVF Conference on Computer Vision and
  Pattern Recognition (CVPR)}, June 2020.

\bibitem{cao2019openpose}
Zhe Cao, Gines Hidalgo, Tomas Simon, Shih-En Wei, and Yaser Sheikh.
\newblock Openpose: Realtime multi-person 2d pose estimation using part
  affinity fields, 2019.

\bibitem{shiftgcn}
Ke Cheng, Yifan Zhang, Xiangyu He, Weihan Chen, Jian Cheng, and Hanqing Lu.
\newblock Skeleton-based action recognition with shift graph convolutional
  network.
\newblock In {\em CVPR 2020}.

\bibitem{haa500}
Jihoon Chung, Cheng hsin Wuu, Hsuan ru Yang, Yu-Wing Tai, and Chi-Keung Tang.
\newblock Haa500: Human-centric atomic action dataset with curated videos.
\newblock In {\em ICCV 2021}.

\bibitem{du2015hierarchical}
Yong Du, Wei Wang, and Liang Wang.
\newblock Hierarchical recurrent neural network for skeleton based action
  recognition.
\newblock In {\em Proceedings of the IEEE conference on computer vision and
  pattern recognition}, pages 1110--1118, 2015.

\bibitem{fang2017rmpe}
Hao-Shu Fang, Shuqin Xie, Yu-Wing Tai, and Cewu Lu.
\newblock {RMPE}: Regional multi-person pose estimation.
\newblock In {\em ICCV}, 2017.

\bibitem{fei2006one}
Li Fei-Fei, Rob Fergus, and Pietro Perona.
\newblock One-shot learning of object categories.
\newblock {\em IEEE transactions on pattern analysis and machine intelligence},
  28(4):594--611, 2006.

\bibitem{article}
Sumaira Ghazal, Umar Khan, Mubasher Saleem, Nasir Rashid, and Javaid Iqbal.
\newblock Human activity recognition using 2d skeleton data and supervised
  machine learning.
\newblock {\em IET Image Processing}, 13, 09 2019.

\bibitem{grauman2021ego4d}
Kristen Grauman, Andrew Westbury, Eugene Byrne, Zachary Chavis, Antonino
  Furnari, Rohit Girdhar, Jackson Hamburger, Hao Jiang, Miao Liu, Xingyu Liu,
  Miguel Martin, Tushar Nagarajan, Ilija Radosavovic, Santhosh~Kumar
  Ramakrishnan, Fiona Ryan, Jayant Sharma, Michael Wray, Mengmeng Xu,
  Eric~Zhongcong Xu, Chen Zhao, Siddhant Bansal, Dhruv Batra, Vincent
  Cartillier, Sean Crane, Tien Do, Morrie Doulaty, Akshay Erapalli, Christoph
  Feichtenhofer, Adriano Fragomeni, Qichen Fu, Christian Fuegen, Abrham
  Gebreselasie, Cristina Gonzalez, James Hillis, Xuhua Huang, Yifei Huang,
  Wenqi Jia, Weslie Khoo, Jachym Kolar, Satwik Kottur, Anurag Kumar, Federico
  Landini, Chao Li, Yanghao Li, Zhenqiang Li, Karttikeya Mangalam, Raghava
  Modhugu, Jonathan Munro, Tullie Murrell, Takumi Nishiyasu, Will Price,
  Paola~Ruiz Puentes, Merey Ramazanova, Leda Sari, Kiran Somasundaram, Audrey
  Southerland, Yusuke Sugano, Ruijie Tao, Minh Vo, Yuchen Wang, Xindi Wu,
  Takuma Yagi, Yunyi Zhu, Pablo Arbelaez, David Crandall, Dima Damen,
  Giovanni~Maria Farinella, Bernard Ghanem, Vamsi~Krishna Ithapu, C.~V.
  Jawahar, Hanbyul Joo, Kris Kitani, Haizhou Li, Richard Newcombe, Aude Oliva,
  Hyun~Soo Park, James~M. Rehg, Yoichi Sato, Jianbo Shi, Mike~Zheng Shou,
  Antonio Torralba, Lorenzo Torresani, Mingfei Yan, and Jitendra Malik.
\newblock Ego4d: Around the world in 3,000 hours of egocentric video, 2021.

\bibitem{Guo_2018_ECCV}
Michelle Guo, Edward Chou, De-An Huang, Shuran Song, Serena Yeung, and Li
  Fei-Fei.
\newblock Neural graph matching networks for fewshot 3d action recognition.
\newblock In {\em ECCV 2018}.

\bibitem{hu2017jointly}
JF Hu, WS Zheng, J Lai, and J Zhang.
\newblock Jointly learning heterogeneous features for rgb-d activity
  recognition.
\newblock {\em IEEE transactions on pattern analysis and machine intelligence},
  39(11):2186--2200, 2017.

\bibitem{hussein2013human}
Mohamed~E Hussein, Marwan Torki, Mohammad~A Gowayyed, and Motaz El-Saban.
\newblock Human action recognition using a temporal hierarchy of covariance
  descriptors on 3d joint locations.
\newblock In {\em Twenty-third international joint conference on artificial
  intelligence}, 2013.

\bibitem{jeon2018constructing}
Yunho Jeon and Junmo Kim.
\newblock Constructing fast network through deconstruction of convolution.
\newblock {\em arXiv preprint arXiv:1806.07370}, 2018.

\bibitem{kay2017kinetics}
Will Kay, Joao Carreira, Karen Simonyan, Brian Zhang, Chloe Hillier, Sudheendra
  Vijayanarasimhan, Fabio Viola, Tim Green, Trevor Back, Paul Natsev, Mustafa
  Suleyman, and Andrew Zisserman.
\newblock The kinetics human action video dataset, 2017.

\bibitem{ke2017new}
Qiuhong Ke, Mohammed Bennamoun, Senjian An, Ferdous Sohel, and Farid Boussaid.
\newblock A new representation of skeleton sequences for 3d action recognition.
\newblock In {\em Proceedings of the IEEE conference on computer vision and
  pattern recognition}, pages 3288--3297, 2017.

\bibitem{kim2017interpretable}
Tae~Soo Kim and Austin Reiter.
\newblock Interpretable 3d human action analysis with temporal convolutional
  networks.
\newblock In {\em 2017 IEEE conference on computer vision and pattern
  recognition workshops (CVPRW)}, pages 1623--1631. IEEE, 2017.

\bibitem{li2017skeleton}
Chao Li, Qiaoyong Zhong, Di Xie, and Shiliang Pu.
\newblock Skeleton-based action recognition with convolutional neural networks.
\newblock In {\em 2017 IEEE International Conference on Multimedia \& Expo
  Workshops (ICMEW)}, pages 597--600. IEEE, 2017.

\bibitem{li2002rapid}
Fei~Fei Li, Rufin VanRullen, Christof Koch, and Pietro Perona.
\newblock Rapid natural scene categorization in the near absence of attention.
\newblock {\em Proceedings of the National Academy of Sciences},
  99(14):9596--9601, 2002.

\bibitem{Li_2020_CVPR}
Shichao Li, Lei Ke, Kevin Pratama, Yu-Wing Tai, Chi-Keung Tang, and Kwang-Ting
  Cheng.
\newblock Cascaded deep monocular 3d human pose estimation with evolutionary
  training data.
\newblock In {\em The IEEE/CVF Conference on Computer Vision and Pattern
  Recognition (CVPR)}, June 2020.

\bibitem{liu2020ntu}
Jun Liu, Amir Shahroudy, Mauricio Perez, Gang Wang, Ling-Yu Duan, and Alex~C
  Kot.
\newblock Ntu rgb+d 120: A large-scale benchmark for 3d human activity
  understanding.
\newblock {\em IEEE Transactions on Pattern Analysis and Machine Intelligence},
  42(10):2684--2701, 2020.

\bibitem{liu2016spatio}
Jun Liu, Amir Shahroudy, Dong Xu, and Gang Wang.
\newblock Spatio-temporal lstm with trust gates for 3d human action
  recognition.
\newblock In {\em European conference on computer vision}, pages 816--833.
  Springer, 2016.

\bibitem{word2vec}
Tomas Mikolov, Ilya Sutskever, Kai Chen, Greg Corrado, and Jeffrey Dean.
\newblock Distributed representations of words and phrases and their
  compositionality.
\newblock In {\em NIPS 2013}.

\bibitem{miller2000learning}
Erik~G Miller, Nicholas~E Matsakis, and Paul~A Viola.
\newblock Learning from one example through shared densities on transforms.
\newblock In {\em Proceedings IEEE Conference on Computer Vision and Pattern
  Recognition. CVPR 2000 (Cat. No. PR00662)}, volume~1, pages 464--471. IEEE,
  2000.

\bibitem{mishra2018generative}
Ashish Mishra, Vinay~Kumar Verma, M~Shiva~Krishna Reddy, S Arulkumar, Piyush
  Rai, and Anurag Mittal.
\newblock A generative approach to zero-shot and few-shot action recognition.
\newblock In {\em 2018 IEEE Winter Conference on Applications of Computer
  Vision (WACV)}, pages 372--380. IEEE, 2018.

\bibitem{7415989}
H. Rahmani, A. Mahmood, D. Huynh, and A. Mian.
\newblock Histogram of oriented principal components for cross-view action
  recognition.
\newblock {\em IEEE Transactions on Pattern Analysis \& Machine Intelligence},
  38(12):2430--2443, dec 2016.

\bibitem{shahroudy2016ntu}
Amir Shahroudy, Jun Liu, Tian-Tsong Ng, and Gang Wang.
\newblock Ntu rgb+d: A large scale dataset for 3d human activity analysis.
\newblock In {\em Proceedings of the IEEE conference on computer vision and
  pattern recognition}, pages 1010--1019, 2016.

\bibitem{smaira2020short}
Lucas Smaira, João Carreira, Eric Noland, Ellen Clancy, Amy Wu, and Andrew
  Zisserman.
\newblock A short note on the kinetics-700-2020 human action dataset, 2020.

\bibitem{tran2015learning}
Du Tran, Lubomir Bourdev, Rob Fergus, Lorenzo Torresani, and Manohar Paluri.
\newblock Learning spatiotemporal features with 3d convolutional networks,
  2015.

\bibitem{Ullah2018ActionRI}
Amin Ullah, Jamil Ahmad, Khan Muhammad, Muhammad Sajjad, and Sung~Wook Baik.
\newblock Action recognition in video sequences using deep bi-directional lstm
  with cnn features.
\newblock {\em IEEE Access}, 6:1155--1166, 2018.

\bibitem{vemulapalli2014human}
Raviteja Vemulapalli, Felipe Arrate, and Rama Chellappa.
\newblock Human action recognition by representing 3d skeletons as points in a
  lie group.
\newblock In {\em CVPR 2014}.

\bibitem{wang2012mining}
Jiang Wang, Zicheng Liu, Ying Wu, and Junsong Yuan.
\newblock Mining actionlet ensemble for action recognition with depth cameras.
\newblock In {\em CVPR 2012}.

\bibitem{wang2016temporal}
Limin Wang, Yuanjun Xiong, Zhe Wang, Yu Qiao, Dahua Lin, Xiaoou Tang, and
  Luc~Van Gool.
\newblock Temporal segment networks: Towards good practices for deep action
  recognition, 2016.

\bibitem{wu2018shift}
Bichen Wu, Alvin Wan, Xiangyu Yue, Peter Jin, Sicheng Zhao, Noah Golmant, Amir
  Gholaminejad, Joseph Gonzalez, and Kurt Keutzer.
\newblock Shift: A zero flop, zero parameter alternative to spatial
  convolutions.
\newblock In {\em Proceedings of the IEEE Conference on Computer Vision and
  Pattern Recognition}, pages 9127--9135, 2018.

\bibitem{xu2018dense}
Baohan Xu, Hao Ye, Yingbin Zheng, Heng Wang, Tianyu Luwang, and Yu-Gang Jiang.
\newblock Dense dilated network for few shot action recognition.
\newblock In {\em Proceedings of the 2018 ACM on International Conference on
  Multimedia Retrieval}, pages 379--387, 2018.

\bibitem{stgcn}
Sijie Yan, Yuanjun Xiong, and Dahua Lin.
\newblock Spatial temporal graph convolutional networks for skeleton-based
  action recognition.
\newblock In {\em Proceedings of the AAAI conference on artificial
  intelligence}, volume~32, 2018.

\bibitem{perminv}
Hongguang Zhang, Li Zhang, Xiaojuan Qi, Hongdong Li, Philip~HS Torr, and Piotr
  Koniusz.
\newblock Few-shot action recognition with permutation-invariant attention.
\newblock In {\em ECCV 2020}.

\bibitem{zhang2017view}
Pengfei Zhang, Cuiling Lan, Junliang Xing, Wenjun Zeng, Jianru Xue, and Nanning
  Zheng.
\newblock View adaptive recurrent neural networks for high performance human
  action recognition from skeleton data.
\newblock In {\em Proceedings of the IEEE International Conference on Computer
  Vision}, pages 2117--2126, 2017.

\bibitem{zhang2020semantics}
Pengfei Zhang, Cuiling Lan, Wenjun Zeng, Junliang Xing, Jianru Xue, and Nanning
  Zheng.
\newblock Semantics-guided neural networks for efficient skeleton-based human
  action recognition.
\newblock In {\em Proceedings of the IEEE Conference on Computer Vision and
  Pattern Recognition}, 2020.

\bibitem{zhong2018shift}
Huasong Zhong, Xianggen Liu, Yihui He, and Yuchun Ma.
\newblock Shift-based primitives for efficient convolutional neural networks.
\newblock {\em arXiv preprint arXiv:1809.08458}, 2018.

\end{thebibliography}
}

\end{document}



\title{HAA4D: Few-Shot Human Atomic Action Recognition via \\ 3D Spatio-Temporal Skeletal Alignment (Supplementary Material)}

\author{Mu-Ruei Tseng$^1$, Abhishek Gupta$^1$, Chi-Keung Tang$^1$, Yu-Wing Tai$^2$  \\
$^1$The Hong Kong University of Science and Technology, $^2$Kuaishou Technology \\
}

\maketitle

\section{Action Classes}
This section provides the complete list of actions contained in our HAA4D dataset. The dataset can be characterized in two parts: \textbf{primary classes} and \textbf{additional classes}, where the former contains 155 actions and the latter contains 145 actions.

\subsection{Primary Classes}
Table \ref{tab:primary} shows actions in the primary classes. Primary classes are actions that contain 20 samples per class. We also include the mean average precision of each class.

\begin{table*}
  \begin{center}
  \resizebox{\textwidth}{!}{%
  \begin{tabular}{|l|l|l|l|l|}
  \hline
1: ALS\_IceBucket\_Challenge (71.65) & 2: CPR (93.86) & 3: abseiling (92.08) & 4: add\_new\_car\_tire (97.45) & 5: adjusting\_glasses (14.61) \\
  \hline
6: air\_drumming (67.30) & 7: air\_guitar (60.32) & 8: air\_hocky (30.83) & 9: alligator\_wrestling (95.37) & 10: applauding (13.73) \\
\hline
11: applying\_cream (1.28) & 12: archery (22.82) & 13: arm\_wave (76.82) & 14: arm\_wrestling (24.19) & 15: atlatl\_throw (64.64) \\ 
\hline
16: axe\_throwing (54.45)&	17: backflip (81.08)&	18: backward\_roll (77.50)&	19: badminton\_overswing (57.07)&	20: badminton\_serve (57.07) \\
\hline
21: badminton\_underswing (64.02)&	22: balancebeam\_flip (33.77)&	23: balancebeam\_jump (65.63)&	24: balancebeam\_rotate (76.87)&	25: balancebeam\_spin (34.47)\\
\hline
26: balancebeam\_walk (73.19)&	27: bandaging (43.44)&	28: baseball\_bunt (96.53)&	29: baseball\_catch\_catcher (79.88)&	30: baseball\_catch\_groundball (91.92)\\
\hline
31: baseball\_pitch (52.28)&	32: baseball\_swing (91.10)&	33: basketball\_dribble (97.64)&	34: basketball\_jabstep (93.37)&	35: basketball\_layup (91.75)\\
\hline
36: basketball\_pass (78.67)&	37: basketball\_shoot (62.81)&	38: battle-rope\_jumping-jack (96.08)&	39: battle-rope\_power-slam (93.65)&	40: battle-rope\_rainbow (98.22)\\
\hline
41: battle-rope\_russian-twist (97.14)&	42: battle-rope\_sideplank (96.29)&	43: battle-rope\_snake (97.66)&	44: battle-rope\_wave (97.42)&	45: beer\_pong\_throw (44.87) \\
\hline
46: bench\_dip (88.79)&	47: bending\_back (76.46)&	48: bowing\_waist (61.65)&	49: brushing\_floor (91.05)&	50: burpee (94.06)\\
\hline
51: climb\_stair (86.19)&	52: fish-hunting\_hold (95.77)&	53: floss\_dance (79.09)&	54: gym\_squat (86.11)&	55: jumping\_jack (82.69)\\
\hline
56: play\_accordian (36.03)&	57: play\_bagpipes (49.74)&	58: play\_bangu (85.08)&	59: play\_banjo (41.72)&	60: play\_castanets (33.98) \\
\hline
61: play\_cello (83.35)&	62: play\_clarinet (20.90)&	63: play\_cornett (15.00)&	64: play\_cymbals (51.04)&	65: play\_diabolo (84.00) \\
\hline
66: play\_doublebass (77.34)&	67: play\_erhu (69.33)&	68: play\_gong (63.88)&	69: play\_grandpiano (81.06)&	70: play\_guitar (70.81)\\
\hline
71: play\_handpan (79.50)&	72: play\_harmonica (34.92)&	73: play\_harp (25.47)&	74: play\_hulahoop (94.84)&	75: play\_hulusi (19.34)\\
\hline
76: play\_kendama (86.82)&	77: play\_leaf-flute (27.48)&	78: play\_lute (66.84)&	79: play\_maracas (39.83)&	80: play\_melodic (41.41)\\
\hline
81: play\_noseflute (15.29)&	82: play\_ocarina (50.90)&	83: play\_otamatone	(12.18)& 84: play\_panpipe (7.74)&	85: play\_piccolo (15.21) \\
\hline
86: play\_recorder (10.25)&	87: play\_sanxian (62.29)&	88: play\_saw (59.23)&	89: play\_saxophone (36.25)&	90: play\_serpent (72.47)\\
\hline
91: play\_sheng (44.04)&	92: play\_sitar (66.23)&	93: play\_suona	(26.76)& 94: play\_tambourine (49.16)&	95: play\_thereminvox (59.91) \\
\hline
96: play\_timpani (76.29)&	97: play\_triangle (24.66)&	98: play\_trombone (12.92)&	99: play\_trumpet (1.77)&	100: play\_ukulele (27.16) \\
\hline
101: play\_viola	(39.31)& 102: play\_violin (32.91)&	103: play\_xylophone (79.20)&	104: play\_yangqin (26.54)&	105: play\_yoyo (69.92) \\
\hline
106: playing\_bass\_drum (41.44)&	107: playing\_cajon\_drum (94.21)&	108: playing\_conga\_drum (71.20)&	109: playing\_nunchucks (84.58)&	110: playing\_rubiks\_cube (31.42)\\
\hline
111: playing\_seesaw (63.97)&	112: playing\_snare\_drum (82.11)& 113: playing\_taiko\_drum (49.75)&	114: pottery\_wheel (90.88)&	115: pouring\_wine (63.42) \\
\hline
116: pull\_ups (84.94)&	117: punching\_sandbag (71.29)&	118: punching\_speedbag (26.99)&	119: push\_car (76.94)&	120: push\_wheelchair (73.89) \\
\hline
121: push\_wheelchair\_alone (86.03)&	122: putting\_scarf\_on (54.48)&	123: quadruped\_hip-extension	(90.14)& 124: racewalk\_walk (65.38)&	125: read\_newspaper (22.82) \\
\hline
126: reading\_book (26.17)&	127: rescue\_breathing (36.16)&	128: sandboarding\_forward (82.46)&	129: side\_lunge (84.16)&	130: situp (88.16) \\
\hline
131: water\_skiing (90.00)&	132: watering\_plants (75.21)&	133: wear\_face\_mask (8.97)&	134: wear\_helmet (29.33)&	135: weightlifting\_hang (96.55) \\
\hline
136: weightlifting\_overhead (98.39)&	137: weightlifting\_stand (97.49)&	138: whipping (96.19)&	139: whistle\_one\_hand (22.97)&	140: whistle\_two\_hands (4.89) \\
\hline
141: workout\_chest-pull (84.69)& 142: workout\_crunch (95.77)&143: yawning	(18.66)&144: yoga\_bridge (98.18)& 145: yoga\_cat (93.47) \\ 
\hline
146: yoga\_dancer (97.69)&	147: yoga\_firefly (96.16)&	148: yoga\_fish (96.51)&	149: yoga\_gate (97.60)& 150: yoga\_locust (90.82) \\
\hline
151: yoga\_lotus (97.93)& 152: yoga\_pigeon (97.67) &153: yoga\_tree (98.45)& 154: yoga\_triangle (98.03)& 155: yoga\_updog (92.47) \\
\hline
  \end{tabular}%
  }
  \end{center}
  \caption{Actions in the primary classes with their mean average precision (mAP). Overall mAP: 63.71.}
  \label{tab:primary}
\end{table*}

\subsection{Additional Classes}
Table \ref{tab:additional} shows actions in the primary classes. Additional classes are actions that contain two samples per class, which are dedicated to one-shot learning. It helps evaluate the performance of the proposed model to differentiate the action when very limited data are presented.

\begin{table*}
  \begin{center}
  \resizebox{\textwidth}{!}{%
  \begin{tabular}{|l|l|l|l|l|}
  \hline
156: balloon\_animal&	157: base\_jumping&	158: baseball\_catch\_flyball&	159: baseball\_run&	160: basketball\_dunk \\
\hline
161: basketball\_hookshot&	162: belly\_dancing&	163: bike\_fall&	164: billiard\_hit&	165: blow\_gun \\
\hline
166: blowdrying\_hair&	167: blowing\_balloon&	168: blowing\_glass&	169: blowing\_gum&	170: blowing\_kisses \\
\hline
171: blowing\_leaf&	172: blowing\_nose&	173: bmx\_jumping&	174: bmx\_riding&	175: bowing\_fullbody \\
\hline
176: bowling&	177: bowls\_throw&	178: breakdancing\_flare&	179: breakdancing\_flip&	180: breakdancing\_rotate \\
\hline
181: breakdancing\_support&	182: brushing\_hair&	183: brushing\_teeth&	184: building\_snowman&	185: burping \\
\hline
186: calfrope\_catch&	187: calfrope\_rope&	188: calfrope\_subdue&	189: canoeing\_slalom&	190: canoeing\_sprint \\
\hline
191: card\_throw&	192: carrying\_with\_head&	193: cartwheeling&	194: cast\_net&	195: chainsaw\_log \\ 
\hline
196: chainsaw\_tree&	197: chalkboard&	198: chewing\_gum&	199: chopping\_meat&	200: chopping\_wood \\ 
\hline
201: cleaning\_mirror&	202: remove\_car\_tire&	203: ride\_bike&	204: ride\_horse&	205: ride\_motorcycle \\
\hline 
206: ride\_scooter&	207: riding\_elephant&	208: riding\_mechanical\_bull&	209: rock\_balancing&	210: rock\_paper\_scissors \\
\hline
211: roller-skating\_backward&	212: roller-skating\_forward&	213: rolling\_snow&	214: rowing\_boat&	215: running\_in\_place \\
\hline
216: running\_on\_four&	217: runway\_walk&	218: sack\_race&	219: salute&	220: screw\_car\_tire \\
\hline
221: scuba\_diving&	222: shake\_cocktail&	223: shaking\_head&	224: shaving\_beard&	225: shoe\_shining \\
\hline
226: shoot\_dance&	227: shooting\_handgun&	228: shooting\_shotgun&	229: shotput\_throw&	230: shoveling\_snow \\
\hline
231: shuffle\_dance&	232: skateboard\_forward&	233: skateboard\_grind&	234: skateboard\_jump&	235: ski\_backflip \\
\hline
236: ski\_cork&	237: ski\_frontflip&	238: ski\_jump\_land&	239: ski\_jump\_midair&	240: ski\_jump\_slide \\
\hline
241: skydiving&	242: sledgehammer\_strike\_down&	243: sling&	244: slingshot&	245: smoking\_exhale \\
\hline
246: smoking\_inhale&	247: snorkeling&	248: snow\_angel&	249: snowboard\_jump&	250: snowboard\_slide \\
\hline
251: soccer\_dribble&	252: soccer\_header&	253: soccer\_save&	254: soccer\_shoot&	255: soccer\_throw \\
\hline
256: softball\_pitch&	257: speed\_stack&	258: speedskating\_forward&	259: spinning\_basketball&	260: spinning\_book \\
\hline
261: spinning\_plate&	262: spitting\_on\_face&	263: split\_leap&	264: spraying\_wall&	265: sprint\_kneel \\
\hline
266: sprint\_run&	267: sprint\_start&	268: squash\_backhand&	269: squash\_forehand&	270: stomping\_grapes \\
\hline
271: stone\_skipping&	272: styling\_hair&	273: surfing&	274: swimming\_backstroke&	275: swimming\_breast\_stroke \\
\hline
276: swimming\_butterfly\_stroke&	277: swimming\_freestyle&	278: swinging\_axe\_on\_a\_tree&	279: sword\_swallowing&	280: taekwondo\_high\_block \\
\hline
281: taekwondo\_kick&	282: taekwondo\_low\_block&	283: taekwondo\_middle\_block&	284: taekwondo\_punch&	285: taichi\_fan \\
\hline
286: taking\_photo\_camera&	287: taking\_selfie&	288: talking\_megaphone&	289: talking\_on\_phone&	290: tap\_dancing \\
\hline
291: tennis\_backhand&	292: tennis\_forehand&	293: tennis\_serve&	294: three\_legged\_race&	295: throw\_boomerang \\
\hline
296: throw\_paper-plane&	297: throwing\_bouquet&	298: tight-rope\_walking&	299: tire\_pull&	300: tire\_sled \\
\hline
  \end{tabular}%
  }
  \end{center}
  \caption{Actions in the additional classes.}
  \label{tab:additional}
\end{table*}

\section{Globally Aligned Skeletons}
Globally aligned skeletons are skeletons that we manually rotated so that all the samples are facing the negative $z$-direction at the start of the action. This adjustment eliminates view variances so that all actions to be performed in the same coordinates space. There are 40 different actions in the primary classes containing globally aligned skeletons. We will provide the list in Table \ref{tab:globally-aligned}.

\begin{table*}
  \begin{center}
  \resizebox{\textwidth}{!}{%
  \begin{tabular}{|l|l|l|l|l|}
  \hline
1: ALS\_IceBucket\_Challenge& 3: abseiling& 4: add\_new\_car\_tire& 5: adjusting\_glasses& 6: air\_drumming \\
\hline
7: air\_guitar& 8: air\_hocky& 9: alligator\_wrestling& 10: applauding& 11: applying\_cream \\
\hline
12: archery& 13: arm\_wave& 14: arm\_wrestling& 15: atlatl\_throw& 16: axe\_throwing \\
\hline
17: backflip& 18: backward\_roll& 19: badminton\_overswing& 20:badminton\_serve& 21: badminton\_underswing \\
\hline
25: balancebeam\_spin& 26: balancebeam\_walk& 27: bandaging& 28: baseball\_bunt& 30: baseball\_catch\_groundball \\
\hline
31: baseball\_pitch& 32: baseball\_swing& 33: basketball\_dribble& 34: basketball\_jabstep& 35: basketball\_layup \\
\hline
36: basketball\_pass& 37: basketball\_shoot& 38: battle-rope\_jumping-jack& 39: battle-rope\_power-slam& 40: battle-rope\_rainbow \\
\hline
41: battle-rope\_russian-twist& 42: battle-rope\_sideplank& 43: battle-rope\_snake& 46: bench\_dip& 50: burpee \\
\hline
  \end{tabular}%
  }
  \end{center}
  \caption{Actions the contain globally aligned skeletons.}
  \label{tab:globally-aligned}
\end{table*}

\section{HAA4D Evaluation Benchmark}
For actions with 20 examples per class, video indexes 0 to 9 are used for training, while videos from 10 to 19 are used for testing. We perform data augmentation on the first ten samples, and among all, videos 8 and 9 are used for validation. For actions that contain only two samples, the one with a smaller index serves as the query, and the other serves as the support.

\section{Training Configuration}
For training the global alignment network, we perform data augmentation by sampling camera views from a 3-frequency subdivision icosahedron. This gives us 92 additional training samples per example. Since there are 400 examples in HAA4D that are provided with globally aligned skeletons, with the help of data augmentation, we have 36,800 examples of training our global alignment network. We split all the data into training and validated with a ratio of 0.8 under two settings: cross-views and cross-actions. We select 73 views on the icosahedron sphere for cross-views and test the rest 19 views to ensure that our network generates predictions decently while encountering unseen views. We also trained our network on cross-views, i.e., 32 out of the 40 classes, to secure that the model is used to generalize different actions. Here are our training environment and configurations in more details:

\begin{itemize}
\item GPU: GeForce GTX TITAN X and GeForce GTX 1080 Ti
\item Epochs: 300
\item Batch size: 64
\item Optimizer: Adam (starting learning rate: 1e-4, weight decay rate: 1e-6)
\item Loss weight: 10:1 (rotation loss : reconstruction loss)
\end{itemize}

\section{Limitations}
Our dataset contains only a few samples per class, which is intended for enhancing its scalability and extension (e.g. 1 for train and 1 for test, see our 5-way 1-shot experiments in main paper), as unlike the 2D counterparts, HAA4D's 3D+T skeletons have more degree of freedom making meaningful data augmentation easy for training on large datasets. In our case, we sample the 3D skeletons from different viewpoints and rotate the skeletons accordingly. We can also use mirroring or combining skeleton samples in the same class to introduce more variation to the dataset. Unlike 2D skeletons that can only have one rotation parameter, the properties of 3D skeletons help better perform data augmentation without breaking the original skeleton structure.

Our few-shot skeleton-based action recognition architecture currently supports only single-person actions. To accommodate actions involving more than one person, we can use a similar technique as ST-GCN, which utilizes the two skeletons with the highest confidence score in the sequences. For actions that have only one subject, we assume them to be all zeros. With this, we can modify our architecture so that instead of having the shape of (n\_ways, n\_shots, n\_segments, n\_encodings) for explicit skeletons encodings, we add one additional dimension so that the skeletons are in the shape of (n\_ways, n\_shots, 2, n\_segments, n\_encodings). We then calculate the distance, respectively. Since there are four skeleton sequences, assuming q\_s1, q\_s2, s\_s1, and s\_s2 all with the shape of (1, n\_segments, n\_encodings), we use the minimum distance between the possible combination ${d(q\_s1,s\_s1) + d(q\_s2,s\_s2), d(q\_s1,s\_s2) + d(q\_s2,s\_s1)}$. Therefore, we can make this adjustment for multi-person interaction, and the rest of the architecture can remain unchanged.